\documentclass[journal]{IEEEtran}
\ifCLASSINFOpdf
   \usepackage[pdftex]{graphicx}
   \DeclareGraphicsExtensions{.pdf,.jpeg,.png}
\else
\fi
\usepackage{array}
\usepackage{graphicx}
\usepackage{color}
\usepackage{amssymb}
\usepackage{bm}
\usepackage{chemformula}
\usepackage{supertabular}
\usepackage{verbatim}
\usepackage{amsmath}
\ifCLASSOPTIONcompsoc
  \usepackage[caption=false,font=normalsize,labelfont=sf,textfont=sf]{subfig}
\else
  \usepackage[caption=false,font=footnotesize]{subfig}
\fi
\usepackage{url}
\usepackage{makecell}

\usepackage{multirow}
\usepackage{mathtools}

\usepackage[tableposition=top]{caption}
\usepackage{listings}
\newcommand\blfootnote[1]{%
  \begingroup
  \renewcommand\thefootnote{}\footnote{#1}%
  \addtocounter{footnote}{-1}%
  \endgroup
}

\begin{document}
\title{Multi-label Classification with Optimal Thresholding for Multi-composition Spectroscopic Analysis}

\author{Luyun~Gan, Brosnan~Yuen and Tao~Lu}
\markboth{}{Gan {\it et al.}:Multi-label Classification with Optimal Thresholding for Multi-composition Spectroscopic Analysis}
\maketitle
\blfootnote{This work is supported in part by the Nature Science and Engineering Research Council of Canada (NSERC) Discovery (Grant No. RGPIN-2015-06515), Defense Threat Reduction Agency (DTRA) Thrust Area 7, Topic G18 (Grant No. GRANT12500317), and Nvidia Corporation TITAN-X GPU grant.(Corresponding author: Tao Lu)}
\blfootnote{L.~Gan, B.~Yuen and T.~Lu  are with the Department of Electrical and Computer Engineering, University of Victoria, EOW 448, 3800 Finnerty Rd., Victoria, British Columbia, V8P 5C2, Canada, (e-mail: \{luyungan,brosnany,taolu\}@uvic.ca)}

\begin{abstract}
 In this paper, we implement multi-label neural networks with optimal thresholding to identify gas species among a multi gas mixture in a cluttered environment. Using infrared absorption spectroscopy and tested on synthesized spectral datasets, our approach outperforms conventional binary relevance - partial least squares discriminant analysis when signal-to-noise ratio and training sample size are sufficient. 
\end{abstract}

\begin{IEEEkeywords}
Multi-label Classification, Infrared Spectroscopy, Supervised learning, Feedforward Neural Networks, Binary Relevance 
\end{IEEEkeywords}

\section{Introduction}
\begin{figure}[h]
    \subfloat[\label{fig:structure_a}]{\includegraphics[width= \columnwidth]{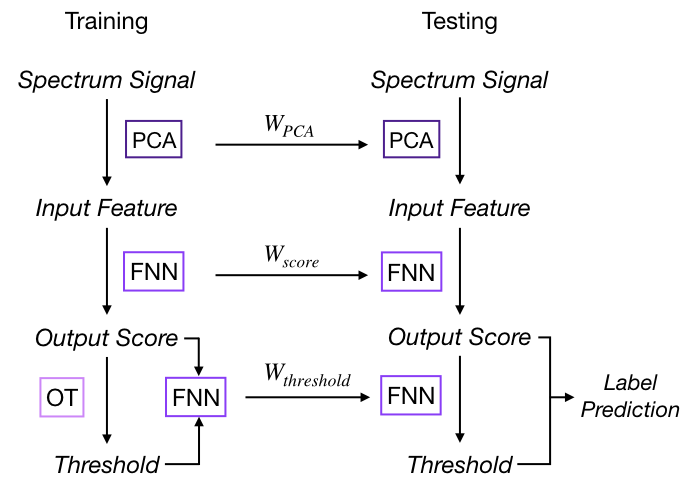}}\\
    \subfloat[\label{fig:0FNN}]{\includegraphics[width= \columnwidth]{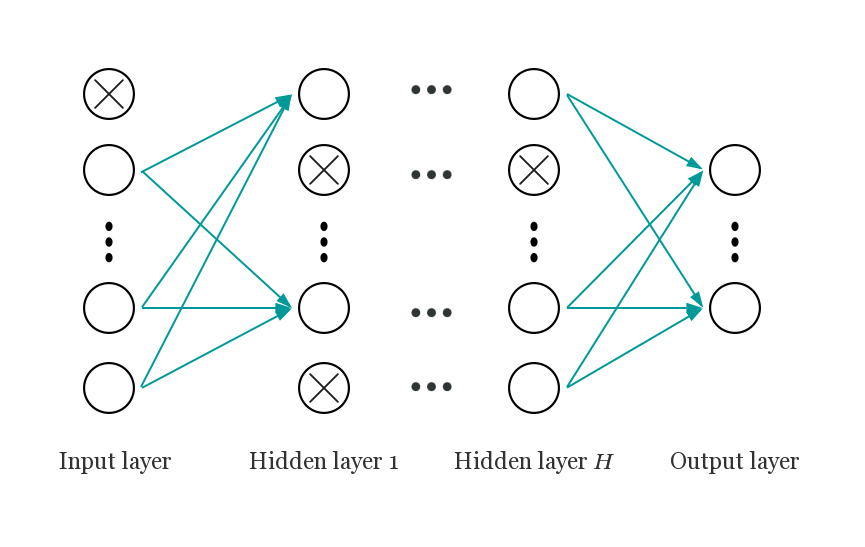}}\\
    \subfloat[\label{fig:structure_b}]{\includegraphics[width= \columnwidth]{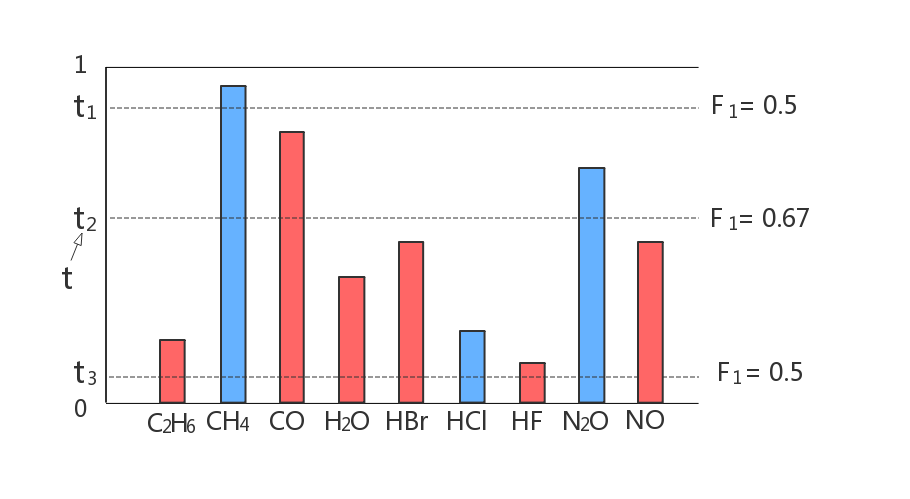}}
    \caption{(a) FNN-OT training and testing procedure. (b) A typical FNN model with dropout. (c) Illustration of optimal thresholding.}
    \label{fig:structure}
\end{figure}
\IEEEPARstart{S}{pectroscopic} analysis sees plural applications in  physics, chemistry, bioinformatics, geophysics, astronomy, etc. It has been widely used for detecting mineral samples~\cite{gallagher2002neural}, gas emission~\cite{jiang2018tdlas} and food volatiles~\cite{dong2018rapid}. Multivariate regression algorithms such as principle component regression~\cite{christy2008real} and partial least squares (PLS)~\cite{wang2018tdlas} are fundamental and popular tools that have been successfully applied to spectroscopic analysis. Non-linear methods,  such as support vector machine~\cite{schumacher2011identification}, genetic programming~\cite{goodacre2003explanatory} and artificial neural networks (ANN)~\cite{gallagher2002neural}, are also adopted to increase prediction accuracy. These algorithms focus on either regression or single-label classification problems.  Using multi-label classification to identify multiple chemical components from the spectrum, is under explored. Unlike multi-classification counterparts that utilizes multiple values of a single label to identify different spectroscopic component, multi-label methods adopt two or more output labels, one for each individual component. Consequently, relations between labels in multi-label tasks can be either independent or correlated.

The development of multi-label classification dates back to 1990s when binary relevance (BR)~\cite{yang1999evaluation} and boosting method~\cite{schapire2000boostexter} were introduced to solve text categorization problems. Significant amount of research was done after that, and the multi-label learning has been prosperous in areas such as natural language processing and image recognition~\cite{tsoumakas2006multi,gibaja2014multi}. Most of the multi-label classification algorithms fall into two basic categories: problem transformation and algorithm adaption. Problem transformation algorithms transform a multi-label problem into one or more single-label problems. After the transformation, existing single-label classifiers can be implemented to make predictions, and the combined outputs will be transformed back into multi-label representations. One of the simplest problem transformation method is BR. It transforms a multi-label problem by splitting it into one binary problem for each label~\cite{godbole2004discriminative,katakis2008multilabel}. Under the assumption of label independence, it ignores the correlations between labels.  If such assumption fails, label powerset (LP) and classifier chains (CC) are known transformation alternatives where LP maps one subset of original labels into one class of the new single label~\cite{tsoumakas2007random} and CC passes label correlation information along a chain of classifiers~\cite{read2009classifier}. In contrast, algorithm adaption methods modify existing single-label classifiers to produce multi-label outputs. For instance, the extensions of decision tree~\cite{clare2001knowledge},  Adaboost~\cite{schapire2000boostexter}, and k-nearest neighbors (KNN)~\cite{zhang2005k} are all designed to deal with multi-label classification problems.  Restricted Boltzman machine~\cite{read2015multi}, feedforward neural network (FNN)~\cite{zhang2006multilabel,nam2014large}, convolutional neural networks (CNN)~\cite{collobert2008unified,gong2013deep}, and recurrent neural networks (RNN)~\cite{wang2016cnn} are employed to characterize label dependency in image processing or find feature representations in text classification. Those adaptive methods can identify multiple labels simultaneously and efficiently without repeatedly trained for sets of labels or chains of classifiers.
\begin{figure}[t!]
    \subfloat[\label{fig:PCA_a}]{\includegraphics[width= \columnwidth]{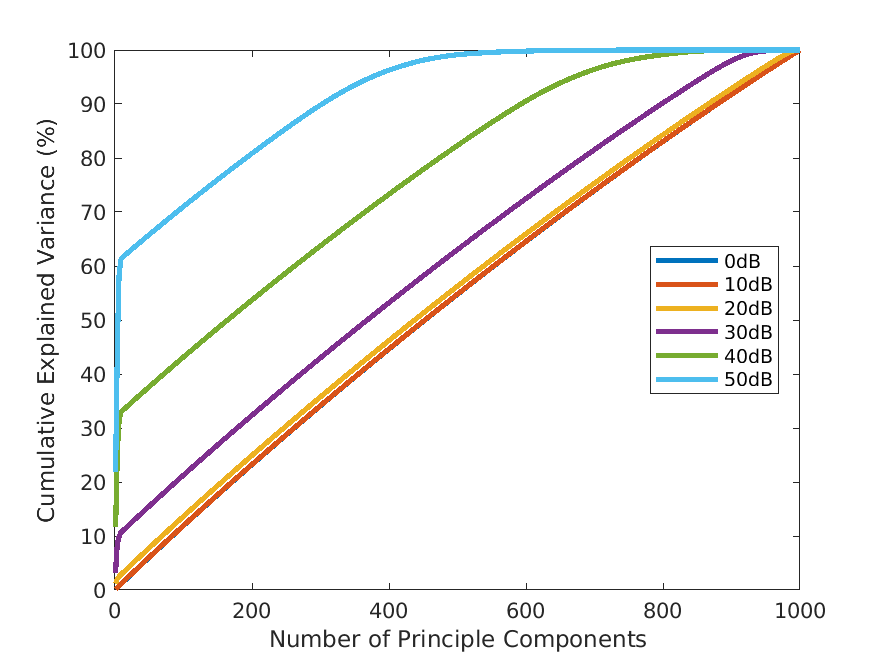}}\\
    \subfloat[\label{fig:PCA_b}]{\includegraphics[width= \columnwidth]{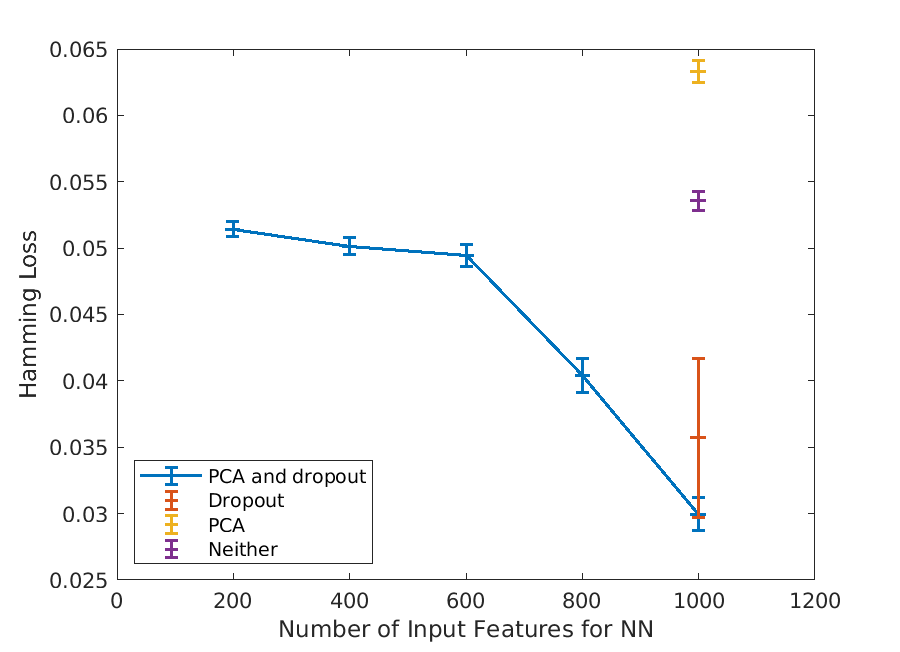}}\\
    \caption{(a) Percentage of cumulative explained variance vs. number of principle components adopted. (b) Comparison of Hamming loss with and without PCA and dropout. }
    \label{fig:PCA}
\end{figure}

Our application of multi-label learning for spectroscopic analysis adopts FNN with optimal thresholding (FNN-OT), which is an adaptive FNN model inspired by~\cite{zhang2006multilabel,nam2014large}. It will be compared with other problem transformation and algorithm adaption models that are extended from  PLS and FNN. In this article, we will train all the models with simulated spectroscopic datasets and compare their results. It will be shown that for most evaluation metrics the adaptive FNN model has the best performance.

\section{Dataset}\label{sec:dataset}
To synthesize the datasets, firstly single gas spectrums of \ch{C2H6}, \ch{CH4}, \ch{CO}, \ch{H2O}, \ch{HBr}, \ch{HCl}, \ch{HF}, \ch{N2O}, and \ch{NO} gasses were selected from the HITRAN~\cite{rothman2013hitran2012} database. The gas spectrums were down sampled to $\mathrm{1,000}$ pixels equally spaced between $\mathrm{1~{\mu}m}$ and $\mathrm{7~{\mu}m}$ wavelengths. Secondly, the gas concentrations were randomly generated from a uniformly distributed probability density function such that the concentration of each gas is uniformly distributed between $\mathrm{0-10~{\mu}M}$.  Thirdly, in real scenarios, gases could be partially correlated. To verify our model under partially correlated components, we introduce highly positive correlation between some gases so that their concentrations retain a pre-set correlation. The generation of uniformly distribute random variables with target correlation matrix will be discussed in Appendix~\ref{appendixA}. Further, in order to test the validity of our classification model, we modify the concentration matrix such that each gas only appears in $50\%$ of the gas mixture samples. Using the concentration matrix, the absorption spectrum of each gas mixture was synthesized using Beer-Lambert law, assuming that the gas mixture was contained in a 10~cm long sensing region and the light source has uniform intensity across the target wavelengths.  Lastly, artificial Gaussian noises with pre-set signal-to-noise ratio (SNR) were added to the light intensity in order to obtain a closer-to-reality spectrum. 
\begin{figure}[ht]
    \centering
    \includegraphics[width=\columnwidth]{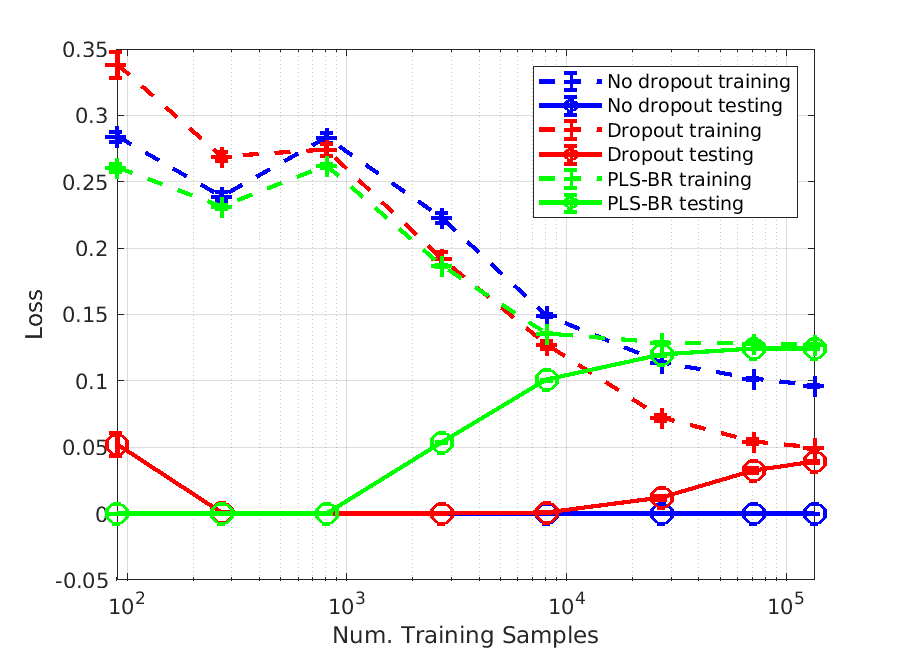}
    \caption{Learning curves of FNN-OT without dropout (blue), with dropout (red) and PLS-BR (green).}
    \label{fig:2LC}
\end{figure}

In this article, we used 12 datasets, each has a pre-set SNR of $\mathrm{0~dB}$, $\mathrm{10~dB}$, $\mathrm{20~dB}$, $\mathrm{30~dB}$, $\mathrm{40~dB}$ and $\mathrm{50~dB}$. For each SNR, we generated two data sets respectively to represent uncorrelated and highly correlated cases. In uncorrelated cases, nine gas labels are mutually independent. In highly correlated cases, nine gases are evenly divided into three subsets. Gas labels within the same subset are highly correlated, and labels from different subsets are independent (Appendix~\ref{appendixA}). 

\section{algorithm}\label{sec:Algorithms}

In single-label learning, a typical approach to classify an instance is to rank the probabilities (or scores) of all classes and choose the class with the highest probability as prediction. For multi-label problems, the same ranking system can be used to compute scores for all labels instead, then a threshold will be determined to assign all labels whose scores are higher than the threshold to the sample. This label score-label prediction framework is the foundation of adapting NN for multi-label learning. In the FNN-OT model, scores of all labels need to be calculated for ranking purpose, and a threshold decision model will be employed to assign a set of labels to the sample in the label prediction step. The whole process of FNN-OT is shown in Fig.~\ref{fig:structure_a}.  Spectrum Signals are firstly pre-processed by principle component analysis (PCA). The output principle components are the input features of an FNN model, which produces one output score for each gas. Output scores will be the input of a following optimal thresholding (OT). For every sample in the training set, its threshold will be determined by OT illustrated in Fig.~\ref{fig:structure_b}. Its mechanism will be explained in Section~\ref{subsec:OT}.  Then the output scores and thresholds are the input and output variables of a new FNN model which will be used to calculate thresholds for testing samples. 

\subsection{Feedforward neural networks}
FNN has outstanding performance with large scale datasets~\cite{nam2014large}. As shown in Fig.~\ref{fig:0FNN}, a typical FNN is formed by an input layer, an output layer and one or more hidden layers in-between. Each layer has a number of active neurons (circles without cross in Fig.~\ref{fig:0FNN}) that use the neuron outputs from previous layer as input and produces output to the neurons in next layer. In our case of multi-label learning, a simple one hidden layer FNN model can achieve a state-of-the-art result with great computational efficiency~\cite{nam2014large}.
To get output score $\textbf{s}$ based on input feature set $\textbf{x}$, our FNN can be written as~\cite{srivastava2014dropout}:

\begin{equation}
\begin{aligned}
    \textbf{h}&=f_h(\textbf{W}^{(1)}\textbf{P}^{(1)}\textbf{x}+\textbf{b}^{(1)})\\
    \textbf{s}&=f_s(\textbf{W}^{(2)}\textbf{P}^{(2)}\textbf{h}+\textbf{b}^{(2)})\\ 
\end{aligned}
\end{equation}

where $\textbf{h}$ is a hidden layer that lies between input and output layer, $f_h$ is the  Rectified  Linear  Units (ReLU) activation function in hidden layer, $f_s$ is the sigmoid function for output layer, and $\textbf{W}^{(1)}$, $\textbf{W}^{(2)}$, $\textbf{b}^{(1)}$, $\textbf{b}^{(2)}$ are the parameters that need to be trained from data.  In our model, the loss function $f_L(\textbf{s},\textbf{y})$ is defined as the cross entropy of label score $\textbf{s}$ and classification target $\textbf{y}$ which can be expressed as:
\begin{equation}
    f_L(\textbf{s},\textbf{y})=-\sum_{i=1}^{L}y_{i}\log(s_{i})+(1-y_{i})\log(1-s_{i})
\end{equation}
where L is the number of labels. 

In our model, we adopted dropout to mitigate overfitting~\cite{srivastava2014dropout}. Dropout is a widely used method for preventing overfitting problems in neural networks. It randomly drops out a percentage of neurons in training, and the weights of remaining neurons will be trained by back-propagation~\cite{srivastava2014dropout}. Retention probability $\boldsymbol{p}=(p_{1}, p_{2})$ is the hyperparameter of dropout that will be tuned for our model. $p_{1}$ and $p_{2}$ are the probabilities of retaining units in input and the hidden layer of the neural network model. Retention probabilities set for the FNN-OT model are the ones that result in minimum losses. 
The dropout is activated by two diagonal matrices of Bernoulli random variables $\textbf{P}^{(1)}$ and $\textbf{P}^{(2)}$ with parameters $p_1$ and $p_2$.  Both parameters are retention probabilities of input and hidden layer for dropout. 

\subsection{Principle component analysis}

In both training and testing, the $\mathrm{1,000}$-pixel absorbance spectra will be pre-processed with principle component analysis (PCA), and the principle components will be the input of the FNN model (\textbf{x}). PCA is a commonly used pre-processing method for spectroscopic datasets. It is conventionally employed to reduce feature dimension by transferring original input variables into a smaller set of uncorrelated principle components (PC) that preserves highest explained variance~\cite{holland2008principal}.  As shown in Fig.~\ref{fig:PCA_a}, at high SNR, PCA is an efficient technique for dimension reduction as only a small number of PCs is sufficient to preserve most of the variances. However, when the SNR drops to below 30~dB, variance of original data is almost evenly projected into PCs. Under such circumstances, PCA will not be efficient for dimension reduction. So, in a preliminary 10-fold test on the SNR=$\mathrm{40~dB}$ dataset, Hamming loss has higher means when number of PCs is less than the number of original pixels (blue line in Fig.~\ref{fig:PCA_b}). However, as shown in the same plot, when PCA is adopted in conjunction with dropout (blue markers), the Hamming loss is significantly reduced compared to the models that only adopts PCA (yellow markers) or dropout (red marker) or neither of them (purple marker). Therefore, in this article, we adopt PCA for all SNRs not only for dimension reduction, but also for Hamming loss reductions.

\subsection{Optimal thresholding}\label{subsec:OT}
Once we obtain the output score $\textbf{s}$ for a specific instance, we need to find a threshold $t_i$ to convert i-th label score $s_i$ in $\textbf{s}$ to i-th label predictions ${\hat y}_i$ in $\hat{\textbf{y}}$. Here, ${\hat y}_i$ can be expressed by an indicator function ${\hat y}_i=1(s_i>t_i)$. That is, for the i-th specific gas component label that has a score higher than $t_i$, the prediction is $1$ and $0$ otherwise, representing the existence/non-existence of that gas component in the spectrum. 

For binary classification problems in single-label learning, the sigmoid activation function of the output layer results in output scores that are between 0 and 1, and those output scores are often interpreted as probabilities of the two possible classes. For each sample in the testing set, its predicted class will be the one with more than 0.5 probability (output score), so the classifier can be viewed as an FNN model with a threshold $t=0.5$. As shown in our result section, mislabelling of extremely low concentration of a specific gas species as absent from the sample occurs more frequently than mislabeling a non-existing gas species as existing in the sample. This results an imbalance between recall and precision. To re-balance recall and precision for higher $\mathrm{F_1}$, adopting an optimal threshold $t$ for each label in each instance is desirable. For samples in the training set, the method of determining $t$ is illustrated in Fig.~\ref{fig:structure_b}. Suppose we have obtained output scores for all nine labels of a gas mixture. Three of them (blue ones) have the ground truth value 1 (gas species exists in the sample), and the rest labels in red are 0 (gas species is absent in the sample). Then we calculate the $\mathrm{F_1}$ scores for the three candidates $t_{1}, t_{2}$ and $t_{3}$ of $t$ (dash lines), and the candidate with the highest $\mathrm{F_1}$ score, which is $t_{2}$ in this example, is the $t$ we need. In our model, we use output scores to calculate the candidates of $t$. For each sample, nine output scores will be formed into an increasing order: $s_{1}\leq s_{2}\leq,...,\leq s_{9}$. Since sigmoid function is used in the output layer, all output scores are between 0 and 1. So the ten threshold candidates will be: 
$$0, \frac{s_{1}+s_{2}}{2},\frac{s_{2}+s_{3}}{2},...,\frac{s_{8}+s_{9}}{2}, 1 $$

In order to systematically get thresholds for all instances in the testing set, we assume that threshold $t$ is determined by the label scores $\textbf{s}$, and their relationship can be recognized by the following FNN model:

\begin{equation}
\begin{aligned}
    \textbf{h}_t&=f_h(\textbf{W}^{(1)}_t\textbf{s}+\textbf{b}^{(1)}_t)\\
    \hat{t}&=\textbf{W}^{(2)}_t\textbf{h}_t+\textbf{b}^{(2)}_t\\ 
\end{aligned}
\end{equation}
\begin{figure}
    \subfloat[\label{fig:1perf_00_a}]{\includegraphics[width= 0.9\columnwidth]{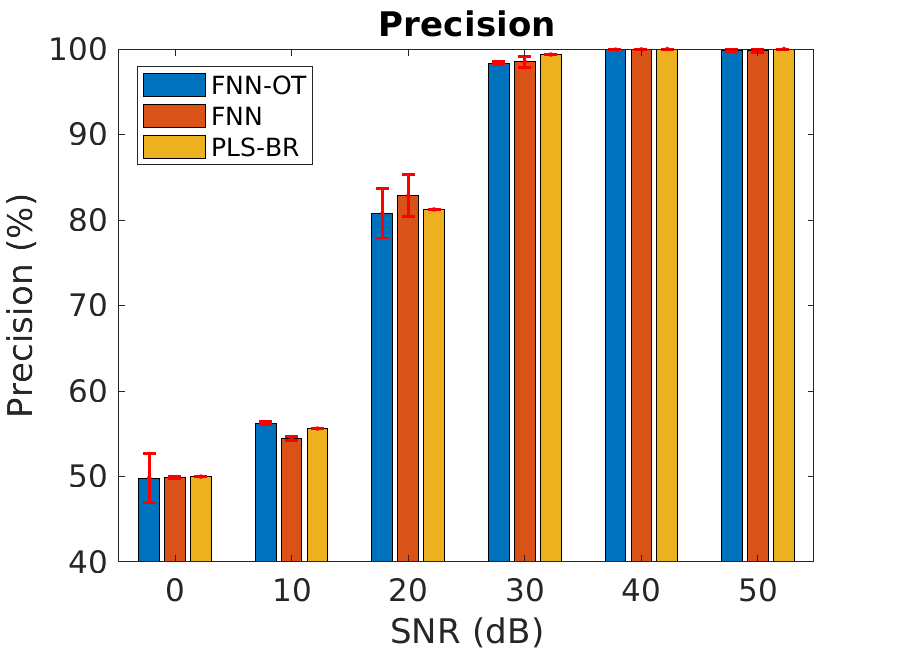}}\\
    \subfloat[\label{fig:1perf_00_b}]{\includegraphics[width= 0.9\columnwidth]{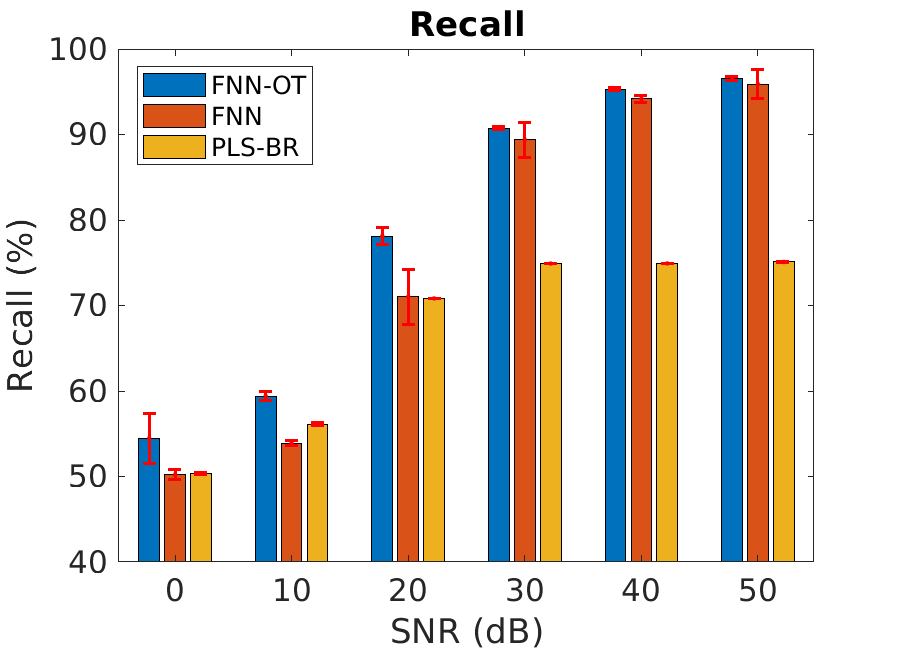}}\\
    \subfloat[\label{fig:1perf_00_c}]{\includegraphics[width= 0.9\columnwidth]{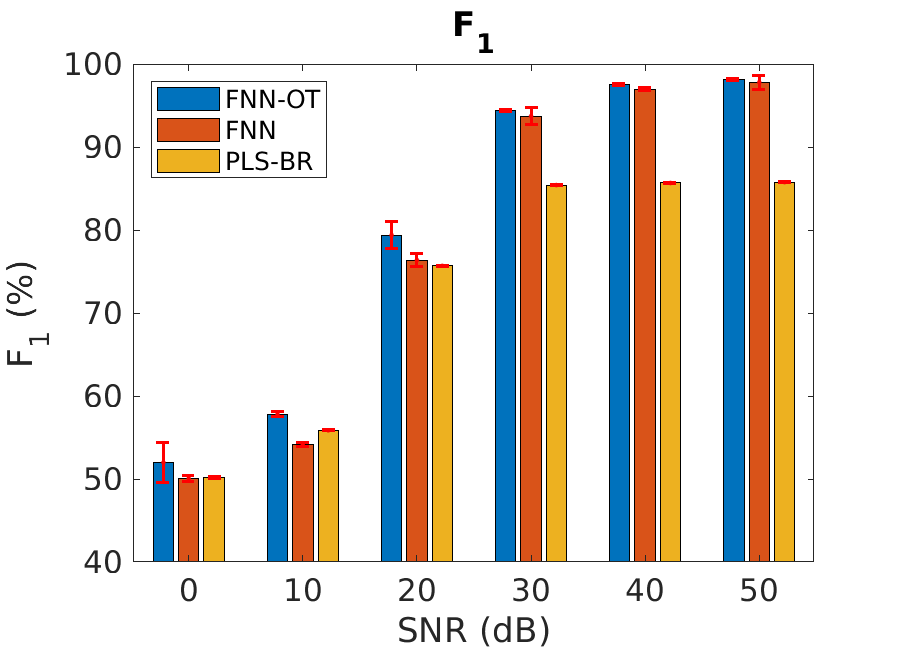}}
    \caption{Micro-averaged (a) precision, (b) recall and (c) $\mathrm{F_1}$ score at different SNRs, assuming all gases are independent.}
    \label{fig:1perf_00}
\end{figure}
\begin{figure*}
    \centering
    \subfloat[\label{fig:3_0db}]{\includegraphics[width=0.33\textwidth]{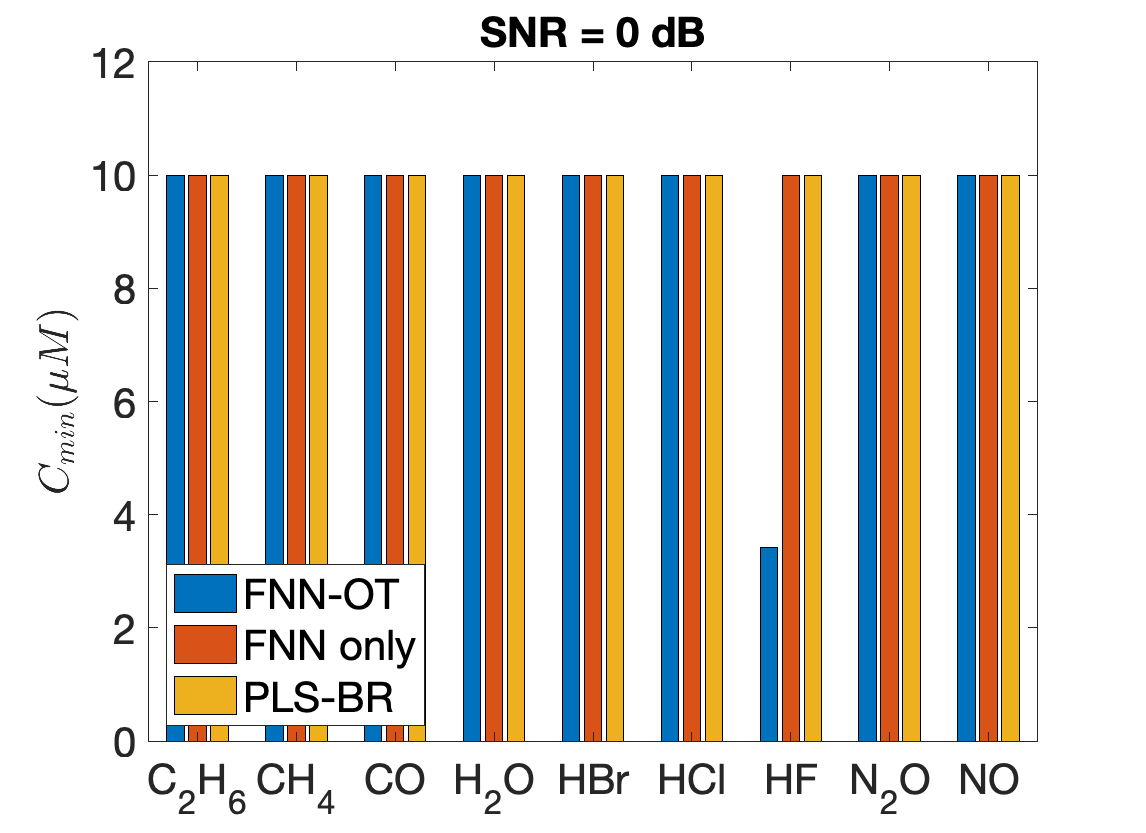}}
    \hfill
    \subfloat[\label{fig:3_10db}]{\includegraphics[width=0.33\textwidth]{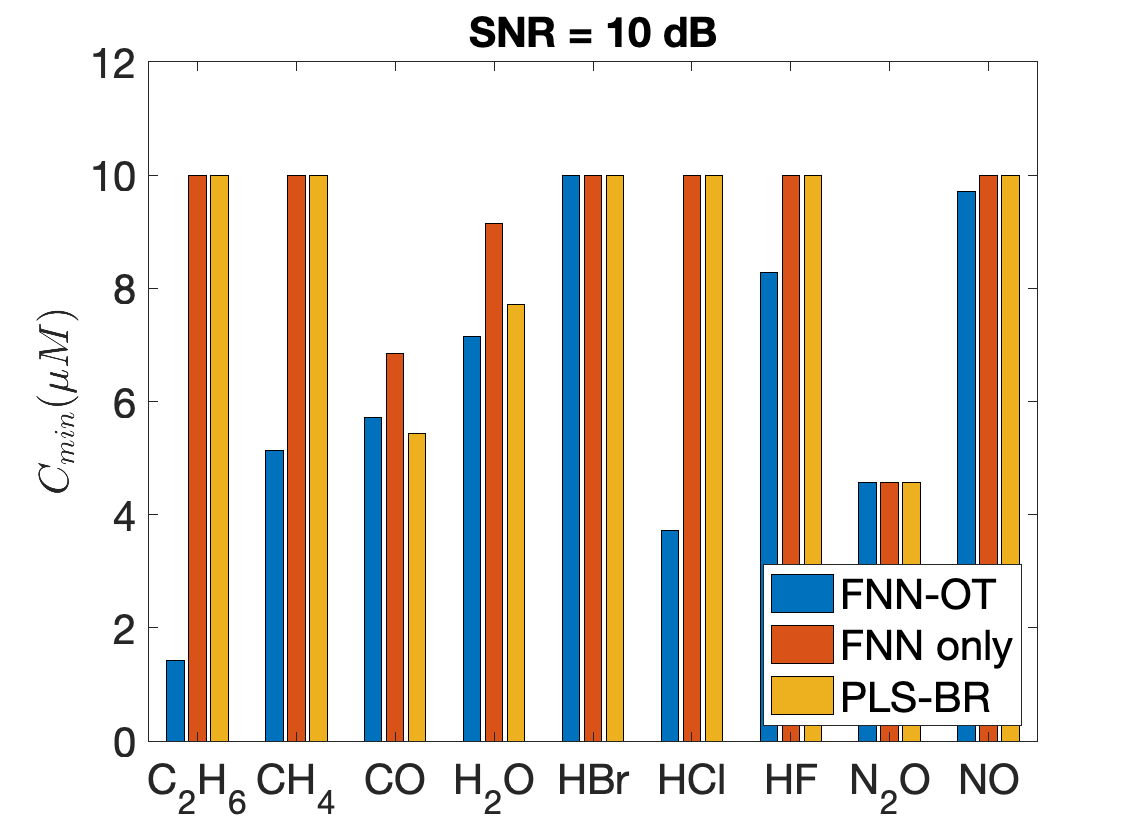}}
    \hfill
    \subfloat[\label{fig:3_20db}]{\includegraphics[width=0.33\textwidth]{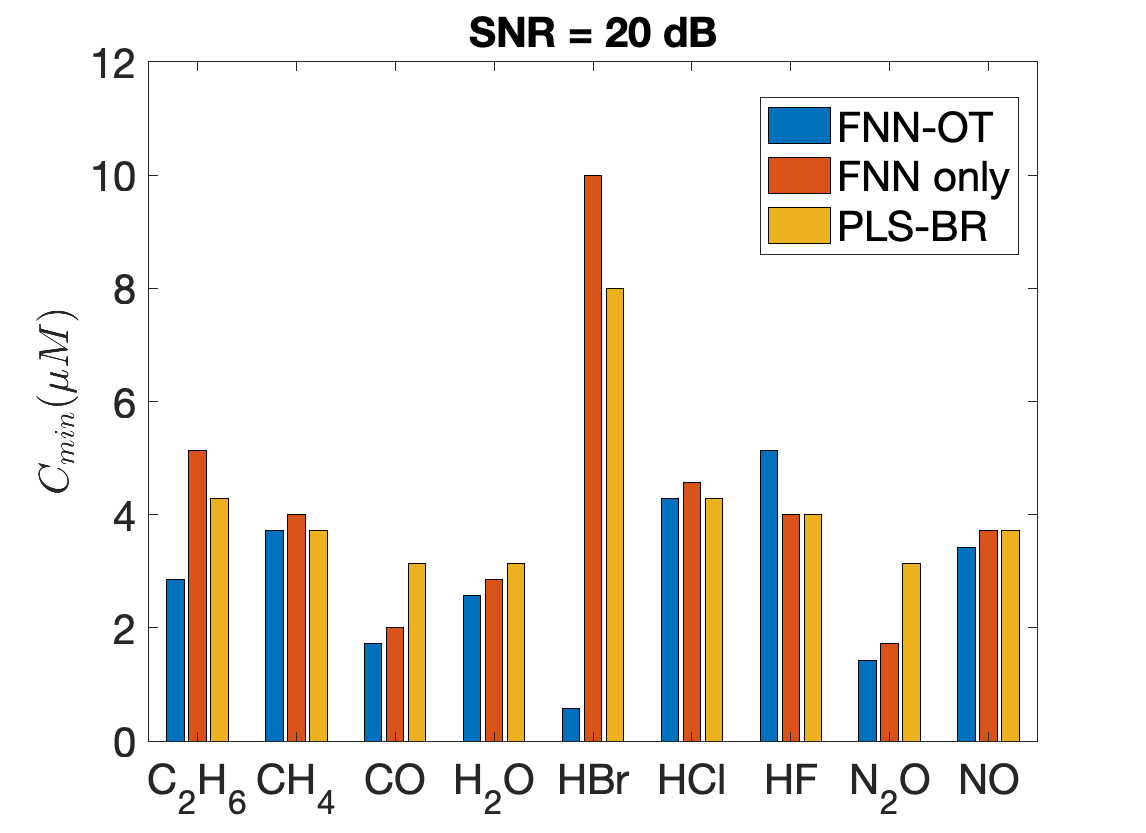}}
    \\
    \subfloat[\label{fig:3_30db}]{\includegraphics[width=0.33\textwidth]{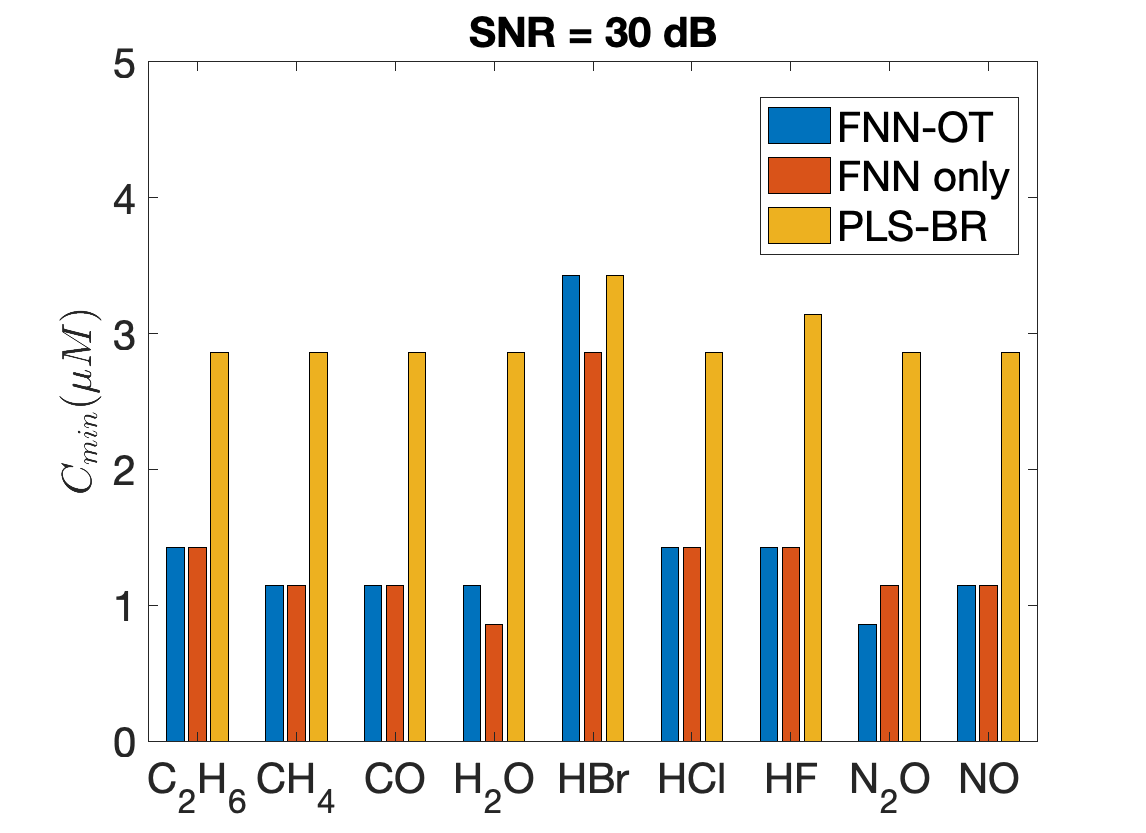}}
    \hfill
    \subfloat[\label{fig:3_40db}]{\includegraphics[width=0.33\textwidth]{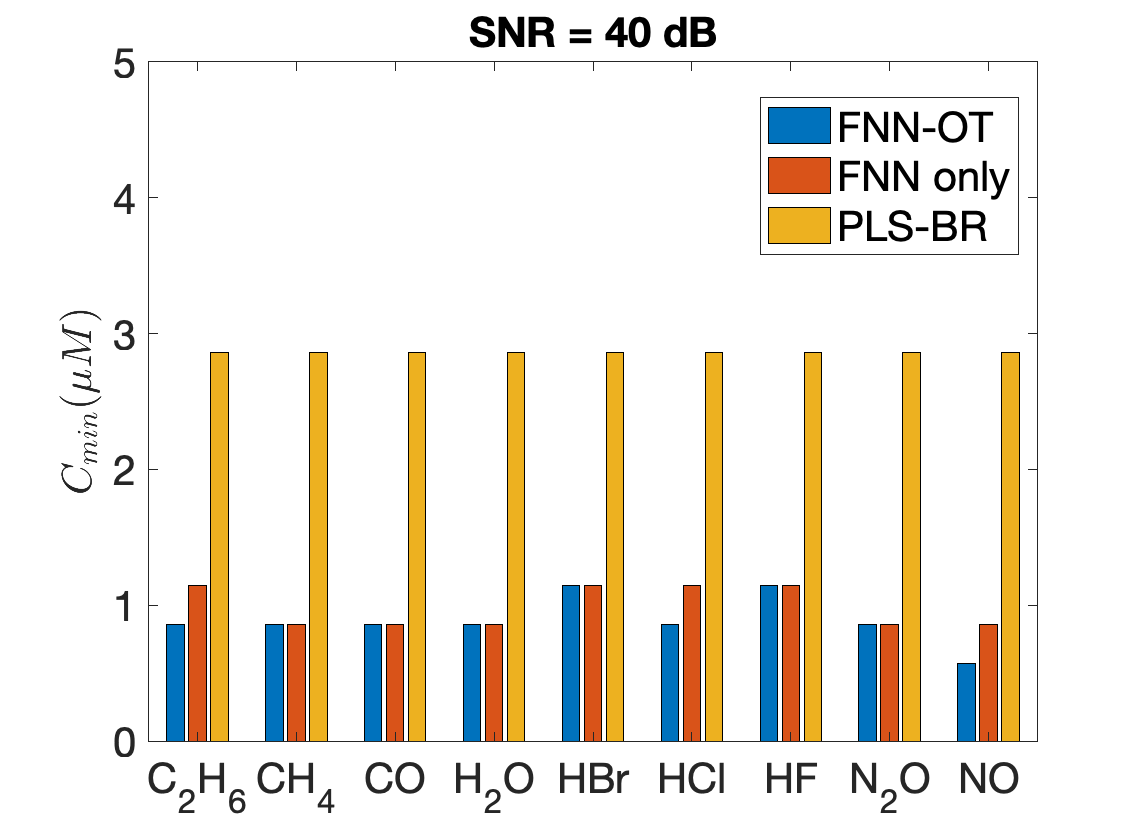}}
    \hfill
    \subfloat[\label{fig:3_50db}]{\includegraphics[width=0.33\textwidth]{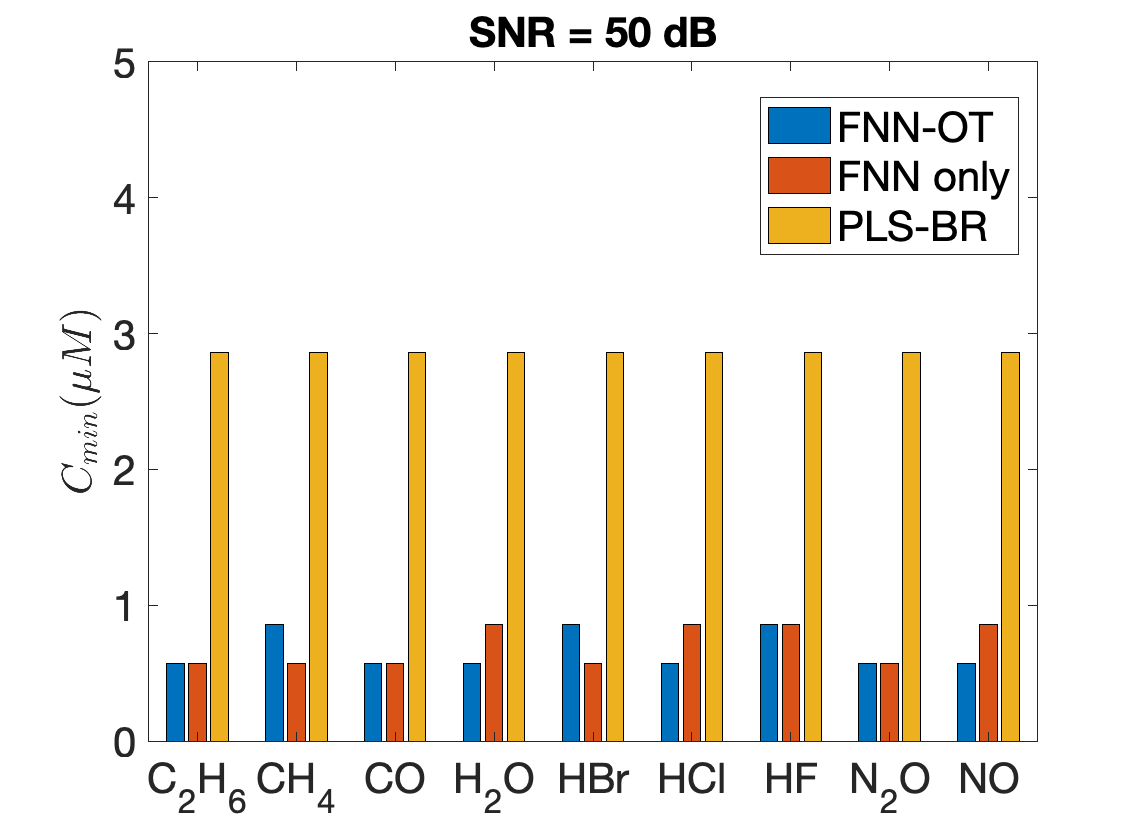}}
    \caption{\label{fig:3mini} Minimum detectable concentration ($C_{min}$) of nine gases.}
\end{figure*}

where $\textbf{h}_t$ is a hidden layer with ReLU activation function $f_h$, and $\textbf{W}^{(1)}_t$, $\textbf{W}^{(2)}_t$,
 $\textbf{b}^{(1)}_t$, $\textbf{b}^{(2)}_t$ are the parameters that need to be estimated. 
We will use instances in the training set to train FNN model, and the loss function is the mean square error between $t$ and $\hat{t}$. 

\subsection{Evaluation metrics}\label{sec:evaluation}
To evaluate our models, we use micro averaged recall, precision and $\mathrm{F_1}$ as our figures of merit.~\cite{tsoumakas2006multi}  In our context, true negative (TN) is the absence of a certain gas that has been correctly predicted in a sample. Similarly, true positive (TP) is the case that an existing gas is marked as present in a sample. False negative (FN) is the case that the classifier fails to identify an existing gas, and false positives (FP) is a false alarm where the classifier identifies a non-existence gas.

\section{Results and Discussions}
\subsection{Hyper parameter tuning}

In our research, we use TensorFlow to implement our FNN-OT and Adam as our optimizer. In first step we tune hyper-parameters such as dropout rate and training sample size of the FNN-OT model with the SNR=$\mathrm{30~dB}$ data set. 

\subsubsection{Dropout}

In order to tune the hyper-parameters for dropout, a grid search has been conducted on retention probabilities $\boldsymbol{p}=(p_{1}, p_{2})$ of input and hidden layers. A typical choice of retention rate is 0.8 for input layer and 0.5 for hidden layer \cite{srivastava2014dropout}, and a preliminary search on our datasets shows that the optimal choice of $\boldsymbol{p}$ is around (0.95, 0.2). 

\subsubsection{Training sample size}
To determine the number of training samples that are sufficient for our models, we plotted the learning curve as shown in Fig.~\ref{fig:2LC}. Here, FNN-OT is compared with PLS-BR (Appendix~\ref{Appendix_PLS}) and FNN with 0.5 threshold. 

As shown in the learning curves plot, we change the number of samples in the training set while keeping the 20,000-sample testing set intact. Both training (solid lines) and testing (dashed lines) Hamming losses are plotted as a function of training samples. At an SNR of $\mathrm{30~dB}$, without dropout (blue markers), our FNN-OT model displays large variance and low bias as the training loss is almost 0 while the testing loss is above 0.1 even when the training sample size is around 100,000. This is a clear indication of overfitting for training samples fewer than 100,000. In contrast, by adopting dropout (red markers), the overfitting issue is solved and both training and testing loss converge to around 0.05 at around 100,000 training samples. In comparison, PLS-BR (green markers) does not display overfitting at the aforementioned sample size. However, the converged training and testing losses are higher $(>\mathrm{0.1})$ than our FNN-OT model with dropout, indicating our model outperforms this conventional technique. Nevertheless, the plot clearly shows that it is sufficient to use around 100,000 samples to train our FNN-OT with dropout model. 

\subsection{Performance comparison of mutually independent gas data}
Parameters of PCA and FNN models will be trained in the 80,000-sample training sets and deployed in the 20,000-sample test sets. 

We first compare our model using the datasets where all gas components are mutually independent. Fig.~\ref{fig:1perf_00}  presents the micro averaged precision, recall and $\mathrm{F_1}$ score at six different SNRs. Expected, all models perform better at higher SNRs.  When SNR is $\mathrm{0~dB}$, all three classifiers failed to identify gases because 0.5 micro-$\mathrm{F_1}$ score is as good as random guess.  Across all SNRs, FNN-OT yields better precision, recall and $\mathrm{F_1}$ than the conventional PLS-BR, clearly indicating it a superior approach for gas identification. Fig.~\ref{fig:1perf_00} also illustrates that all three models display higher values of precision than recall. This is due to the fact that most mislabelling occurs when a gas species' concentration is too low to produce detectable signal above noise background, all model will mistakenly predict the absence of that gas and produce a FN. However, as evident from Fig.~\ref{fig:1perf_00}b, selecting optimal threshold will significantly reduce the occurrence of FN and increase recall without significantly reducing the precision, resulting a better $\mathrm{F_1}$ score. This clearly justifies the necessity of adopt FNN-OT.

The advantages of FNN-OT are further confirmed by comparing minimum detectable concentrations of nine gases in Fig.~\ref{fig:3mini}. As shown, both FNN-OT and FNN consistently show lower minimum detectable concentration at all SNRs while in general FNN-OT outperforms FNN. 

\subsection{Performance comparison for highly correlated gas data}
\begin{figure}[t!]
    \subfloat[\label{fig:4perf_00_a}]{\includegraphics[width= 0.9\columnwidth]{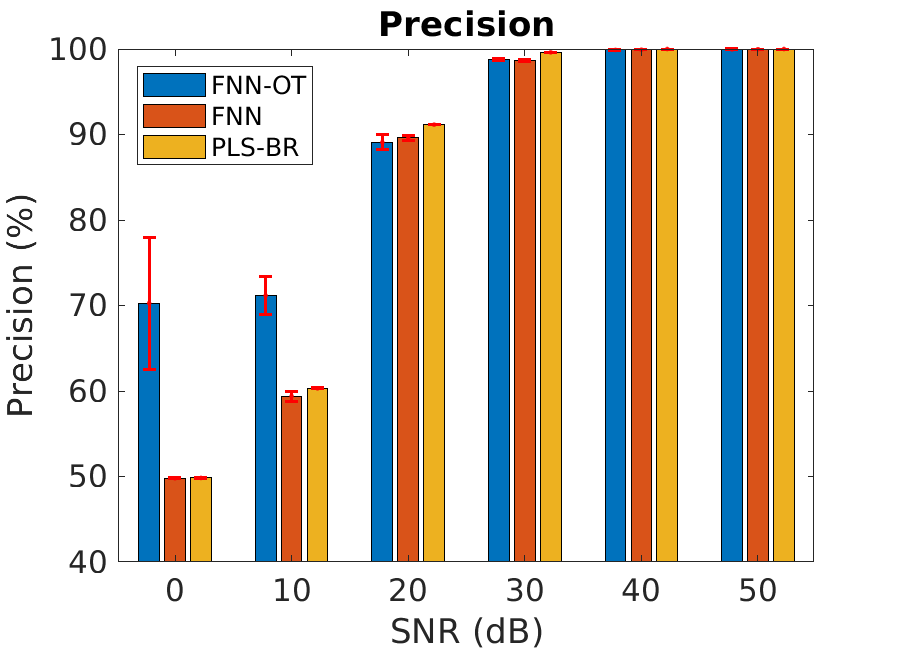}}\\
    \subfloat[\label{fig:4perf_00_b}]{\includegraphics[width= 0.9\columnwidth]{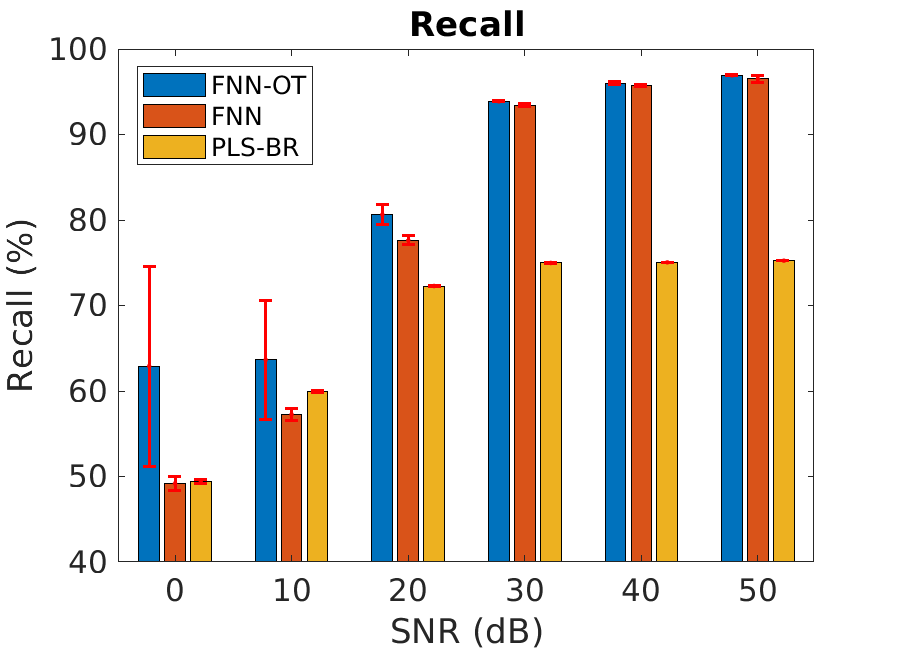}}\\
    \subfloat[\label{fig:4perf_00_c}]{\includegraphics[width= 0.9\columnwidth]{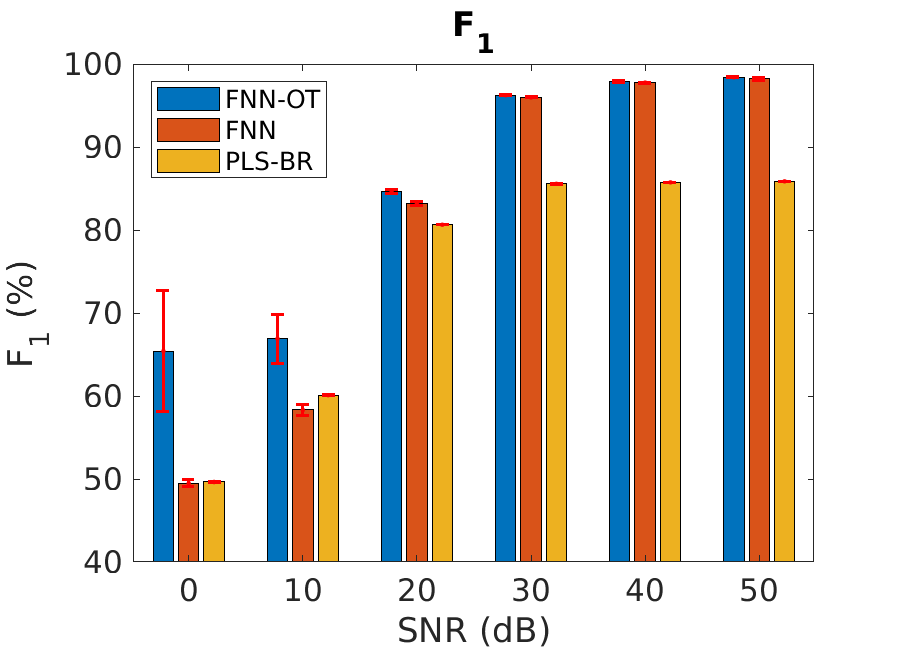}}
    \caption{Micro-averaged (a) precision, (b) recall and (c) $\mathrm{F_1}$ score at different SNRs, assuming some of the gases are highly correlated.}
    \label{fig:4perf_00}
\end{figure}

We further apply our models to the cases when the gases are correlated. As shown in Fig.~\ref{fig:4perf_00}, when SNR is above $\mathrm{20~dB}$, performance of the 3 models is similar to the uncorrelated case and FNN-OT outperforms. Further at SNR=$\mathrm{0~dB}$ or $\mathrm{10~dB}$, FNN-OT significantly outperforms the other 2 models and its own results of the uncorrelated case due to the fact that FNN-OT can collaboratively identify gas species through organize their correlation while FNN and PLS-BR are not capable of. 

\section{Conclusions}

In conclusion, by selecting optimal thresholds, FNN-OT outperforms conventional PLS-BR and FNN in two aspects. FNN-OT can dynamically select a threshold to reduce FN events. In addition, FNN-OT is capable of utilizing correlation among the components to enhance its classification capability. Both of these unique features make FNN-OT a favorable choice for spectroscopic analysis in cluttered environments. 

\appendices
\section{GENERATION OF CORRELATED UNIFORMLY DISTRIBUTED
RANDOM VARIABLES}\label{appendixA}
\begin{figure*}[ht!]
    \centering
    \includegraphics[width=7in]{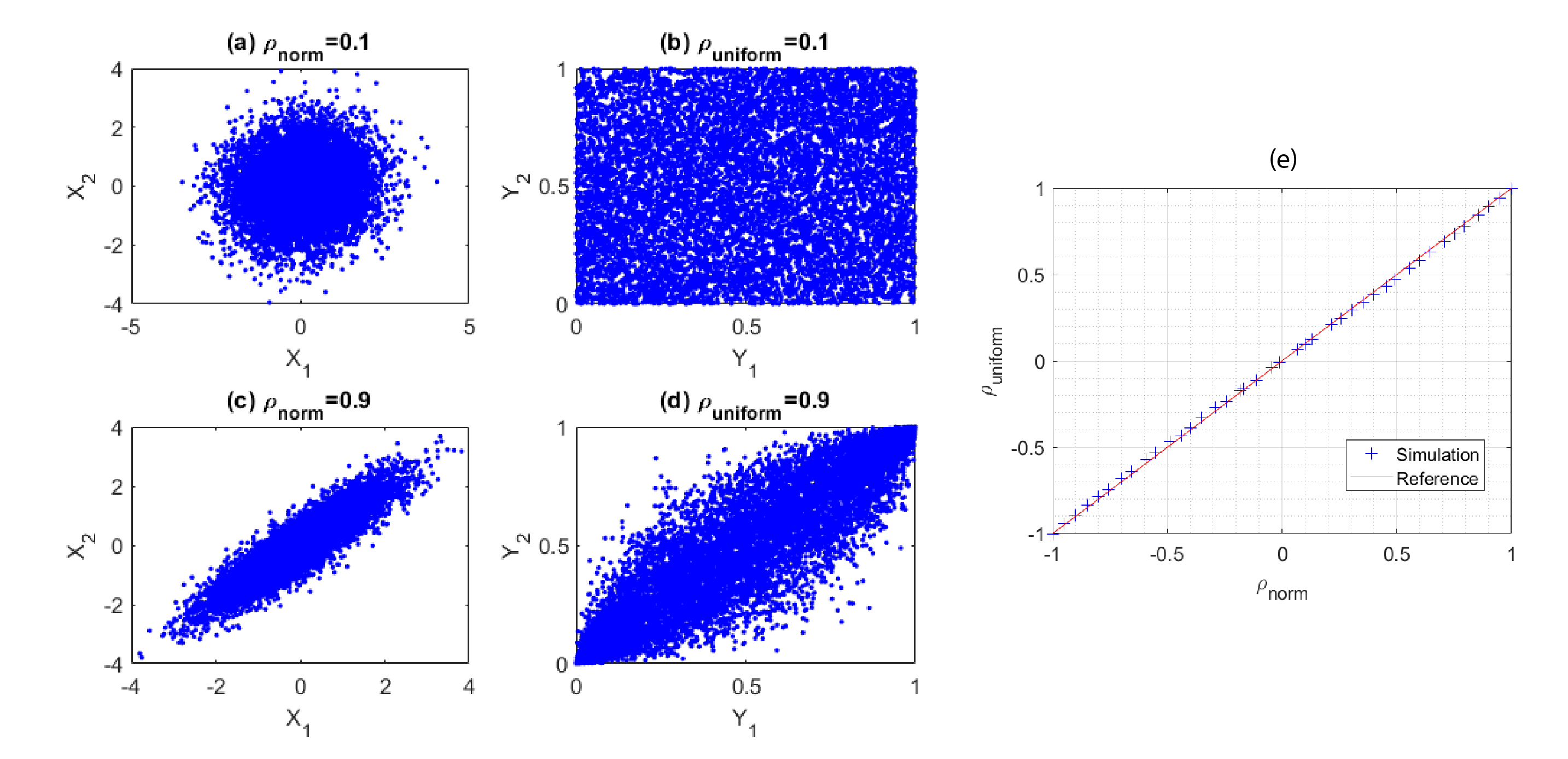}
    \caption{Joint distribution of two normal distributed random variables $X_1$ and $X_2$ with correlation coefficient (a) $\rho_{norm}=0.1$ and (b) $\rho_{norm}=0.9$. The joint distribution of transformed uniform distributed random variables $Y_1$ and $Y_2$ are plotted in (c) and (d) respectively with correlation retained. (e) The correlation coefficients of transformed random variables vs. the coefficients of the original normal distributed random variables. }
    \label{fig_appendixA}
\end{figure*}
To test our models with highly correlated gas labels, we construct a correlation matrix of all nine gases. To simplify our model, we evenly divided nine gases into three subsets and generated highly correlated uniformly distributed random concentrations of the three gases in each subset. Firstly, we generated covariance matrix $\Sigma$ of the nine variables, which has to be symmetric positive semi-definite. Let

\begin{equation}
    \boldsymbol{L}=
    \begin{bmatrix}
    100\boldsymbol{L_{11}} & \boldsymbol{L_{12}} & \boldsymbol{L_{13}} \\
    \boldsymbol{L_{21}} & 100\boldsymbol{L_{22}} & \boldsymbol{L_{23}} \\
    \boldsymbol{L_{31}} & \boldsymbol{L_{32}} & 100\boldsymbol{L_{33}}
    \end{bmatrix}
\end{equation}
where $\boldsymbol{L_{ij}}$ are $3 \times 3$ random matrices with element values uniformly distributed between $(0,1)$. Then $\Sigma=\boldsymbol{LL}^{T}$ will be symmetric positive semi-definite. With the covariance matrix, one may easily obtain the corresponding multivariate normal random numbers $X_i$ through, e.g. MATLAB's mvnrnd command. To generate uniformly distributed random numbers $Y_i$ from the above multivariate normal random numbers $X_i$, we used the approach in~\cite{Allred_Partially}. The procedure  is as follows: define $x_i^j, (j=1,\ldots,N_i)$ as the $j-th$ random number of $X_i, (i=1,2)$. $N_i$ is the total number of samples in random variable $X_i$. First compute the cumulative distributed function $P_{cdf}^i$ of $X_i$ according to
\begin{equation}
    P_{cdf}^i(x)=\frac{1}{N_i}\Sigma_1^{N_i}1(x_i^j<x)
\end{equation}
Here $1(x_i^j<x)$ is an indicator function that returns $1$ if the condition in the bracket holds and $0$ otherwise. Consequently, the uniformly distributed random variable $Y_i, (i=1,2)$ can be easily constructed accordingly to
\begin{equation}
    Y_i=P_{cdf}^i(X_i)
\end{equation}

Fig.~\ref{fig_appendixA} clearly shows the validity of the procedure. Here, joint distribution of two partially correlated normal distributed random variables $X_1$ and $X_2$ with correlation coefficients 0.1 and 0.9 are plotted in subplots (a) and (c) respectively. The distribution of the corresponding transformed uniformly distributed random variables $Y_1$ and $Y_2$ are shown in subplots (b) and (d), with correlation coefficient values retained after transformation. Fig.~\ref{fig_appendixA}(e) further plot the correlation coefficients of $Y$ vs. the coefficients of $X$. As shows, the transformed correlation coefficients are almost identical to the coefficients of their original pair. 

\section{Partial least square method}\label{Appendix_PLS}
Our model compares with conventional PLS-BR. PLS-BR is a multi-label classifier adapted from PLS. It utilizes BR to split the multi-label task into several single-label classification problems. BR decomposes the learning of output labels into a set of binary classification tasks, one per label, where each single model is learned independently, using only the information of that particular label and ignoring the information of all other labels\cite{Luaces2012}. It has various advantages such as the base learner can be selected from any of the binary learning methods, and also the complexity is linear with the number of labels. Apart from this, it can also optimize several loss functions. The main disadvantage of BR is that it assumes that all labels are independent and ignores the correlations between them. 

PLS is a widely used quantitative technique in advanced spectral analysis~\cite{madden2009machine}. In order to predict output $Y$ from feature $X$, PLS describes the common structure of $X$ and $Y$ by combining PCA and multivariate regression.~\cite{geladi1986partial} Similar to PCA, PLS decomposes $\boldsymbol{X}$ and $\boldsymbol{Y}$ as follows:

\begin{equation}
\begin{split}
    \boldsymbol{X}=\boldsymbol{TP}^{T}\\
    \boldsymbol{Y}=\boldsymbol{UQ}^{T}    
\end{split}
\end{equation}

Where $\boldsymbol{T}$ and $\boldsymbol{U}$ are projections of $\boldsymbol{X}$ and $\boldsymbol{Y}$, $\boldsymbol{P}^{T}$ and $\boldsymbol{Q}^{T}$ are transpose of orthogonal loading matrices. Then regression of $\boldsymbol{T}$ and $\boldsymbol{U}$ will be performed following the standard multivariate regression procedure.

PLS itself is not designed for classification, so an extension of PLS called PLS-DA (Partial Least Squares - Discriminant Analysis) is adopted to classify categorical outputs. PLS-DA has been successfully used to classify milk and lubricant based on spectroscopic data sets in \cite{elbassbasi2010classification}, \cite{hirri2013classification}, and \cite{hirri2016ftir}. In binary classification ($y=0$ or $1$) cases, PLS-DA creates two dummy variables $y_1 (y=0)$ and $y_2 (y=1)$ for the $y $ label, and then calculates the PLS regression scores for $y_1$ and $ y_2$. If $y_1$ has higher score, $y$ is classified as 0. Otherwise the prediction class of $y$ is 1.

\section*{Acknowledgment}
  The authors would like to thank Prof. Kerry Vahala and Prof. Qiang Lin for their helpful discussions.

\ifCLASSOPTIONcaptionsoff
  \newpage
\fi


\begin{thebibliography}{10}
\providecommand{\url}[1]{#1}
\csname url@samestyle\endcsname
\providecommand{\newblock}{\relax}
\providecommand{\bibinfo}[2]{#2}
\providecommand{\BIBentrySTDinterwordspacing}{\spaceskip=0pt\relax}
\providecommand{\BIBentryALTinterwordstretchfactor}{4}
\providecommand{\BIBentryALTinterwordspacing}{\spaceskip=\fontdimen2\font plus
\BIBentryALTinterwordstretchfactor\fontdimen3\font minus
  \fontdimen4\font\relax}
\providecommand{\BIBforeignlanguage}[2]{{%
\expandafter\ifx\csname l@#1\endcsname\relax
\typeout{** WARNING: IEEEtran.bst: No hyphenation pattern has been}%
\typeout{** loaded for the language `#1'. Using the pattern for}%
\typeout{** the default language instead.}%
\else
\language=\csname l@#1\endcsname
\fi
#2}}
\providecommand{\BIBdecl}{\relax}
\BIBdecl

\bibitem{gallagher2002neural}
M.~Gallagher and P.~Deacon, ``Neural networks and the classification of
  mineralogical samples using {X}-ray spectra,'' in \emph{Proceedings of the 9th International Conference on Neural Information Processing. ICONIP'02.}, vol.~5. pp.
  2683--2687. IEEE, 2002, 

\bibitem{jiang2018tdlas}
J.~Jiang, M.~Zhao, G.-M. Ma, H.-T. Song, C.-R. Li, X.~Han, and C.~Zhang,
  ``TDLAS-based detection of dissolved methane in power transformer oil and
  field application,'' \emph{IEEE Sensors Journal}, vol.~18, no.~6, pp.
  2318--2325, 2018.

\bibitem{dong2018rapid}
D.~Dong, L.~Jiao, C.~Li, and C.~Zhao, ``Rapid and real-time analysis of
  volatile compounds released from food using infrared and laser
  spectroscopy,'' \emph{TrAC Trends in Analytical Chemistry}, 2018.

\bibitem{christy2008real}
C.~D. Christy, ``Real-time measurement of soil attributes using on-the-go near
  infrared reflectance spectroscopy,'' \emph{Computers and Electronics in
  Agriculture}, vol.~61, no.~1, pp. 10--19, 2008.

\bibitem{wang2018tdlas}
Y.~Wang, Y.~Wei, T.~Liu, T.~Sun, and K.~T. Grattan, ``TDLAS detection of
  propane/butane gas mixture by using reference gas absorption cells and
  partial least square approach,'' \emph{IEEE Sensors Journal}, vol.~18,
  no.~20, pp. 8587--8596.

\bibitem{schumacher2011identification}
W.~Schumacher, M.~K{\"u}hnert, P.~R{\"o}sch, and J.~Popp, ``Identification and
  classification of organic and inorganic components of particulate matter via
  raman spectroscopy and chemometric approaches,'' \emph{Journal of Raman
  Spectroscopy}, vol.~42, no.~3, pp. 383--392, 2011.

\bibitem{goodacre2003explanatory}
R.~Goodacre, ``Explanatory analysis of spectroscopic data using machine
  learning of simple, interpretable rules,'' \emph{Vibrational Spectroscopy},
  vol.~32, no.~1, pp. 33--45, 2003.

\bibitem{yang1999evaluation}
Y.~Yang, ``An evaluation of statistical approaches to text categorization,''
  \emph{Information retrieval}, vol.~1, no. 1-2, pp. 69--90, 1999.

\bibitem{schapire2000boostexter}
R.~E. Schapire and Y.~Singer, ``Boostexter: A boosting-based system for text
  categorization,'' \emph{Machine Learning}, vol.~39, no. 2-3, pp. 135--168,
  2000.

\bibitem{tsoumakas2006multi}
G.~Tsoumakas and I.~Katakis, ``Multi-label classification: An overview,''
  \emph{International Journal of Data Warehousing and Mining}, vol.~3, no.~3,
  2006.

\bibitem{gibaja2014multi}
E.~Gibaja and S.~Ventura, ``Multi-label learning: A review of the state of the
  art and ongoing research,'' \emph{Wiley Interdisciplinary Reviews: Data
  Mining and Knowledge Discovery}, vol.~4, no.~6, pp. 411--444, 2014.

\bibitem{godbole2004discriminative}
S.~Godbole and S.~Sarawagi, ``Discriminative methods for multi-labeled
  classification,'' \emph{Advances in Knowledge Discovery and Data Mining}, pp.
  22--30, 2004.

\bibitem{katakis2008multilabel}
I.~Katakis, G.~Tsoumakas, and I.~Vlahavas, ``Multilabel text classification for
  automated tag suggestion,'' \emph{ECML PKDD Discovery Challenge}, vol.~75,
  2008.

\bibitem{tsoumakas2007random}
G.~Tsoumakas and I.~Vlahavas, ``Random k-labelsets: An ensemble method for
  multilabel classification,'' in \emph{European Conference on Machine
  Learning}.\hskip 1em plus 0.5em minus 0.4em\relax Springer, pp.
  406--417, 2007.

\bibitem{read2009classifier}
J.~Read, B.~Pfahringer, G.~Holmes, and E.~Frank, ``Classifier chains for
  multi-label classification,'' \emph{Machine Learning and Knowledge Discovery
  in Databases}, pp. 254--269, 2009.

\bibitem{clare2001knowledge}
A.~Clare and R.~D. King, ``Knowledge discovery in multi-label phenotype data,''
  in \emph{European Conference on Principles of Data Mining and Knowledge
  Discovery}.\hskip 1em plus 0.5em minus 0.4em\relax Springer, pp.
  42--53, 2001.

\bibitem{zhang2005k}
M.-L. Zhang and Z.-H. Zhou, ``A k-nearest neighbor based algorithm for
  multi-label classification,'' in \emph{2005 IEEE
  International Conference on Granular Computing}, vol.~2.
   pp. 718--721, 2005.

\bibitem{read2015multi}
J.~Read and J.~Hollm{\'e}n, ``Multi-label classification using labels as hidden
  nodes,'' \emph{arXiv preprint arXiv:1503.09022}, 2015.

\bibitem{zhang2006multilabel}
M.-L. Zhang and Z.-H. Zhou, ``Multilabel neural networks with applications to
  functional genomics and text categorization,'' \emph{IEEE Transactions on
  Knowledge and Data Engineering}, vol.~18, no.~10, pp. 1338--1351, 2006.

\bibitem{nam2014large}
J.~Nam, J.~Kim, E.~L. Menc{\'\i}a, I.~Gurevych, and J.~F{\"u}rnkranz,
  ``Large-scale multi-label text classification-revisiting neural networks,''
  in \emph{Joint European Conference on Machine Learning and Knowledge
  Discovery in Databases}.\hskip 1em plus 0.5em minus 0.4em\relax Springer,
   pp. 437--452, 2014.

\bibitem{collobert2008unified}
R.~Collobert and J.~Weston, ``A unified architecture for natural language
  processing: Deep neural networks with multitask learning,'' in
  \emph{Proceedings of the 25th International Conference on Machine
  Learning}.\hskip 1em plus 0.5em minus 0.4em\relax ACM, pp. 160--167, 2008.

\bibitem{gong2013deep}
Y.~Gong, Y.~Jia, T.~Leung, A.~Toshev, and S.~Ioffe, ``Deep convolutional
  ranking for multilabel image annotation,'' \emph{arXiv preprint
  arXiv:1312.4894}, 2013.

\bibitem{wang2016cnn}
J.~Wang, Y.~Yang, J.~Mao, Z.~Huang, C.~Huang, and W.~Xu, ``{CNN-RNN}: A unified
  framework for multi-label image classification,'' in \emph{Proceedings of the
  IEEE Conference on Computer Vision and Pattern Recognition}, pp.
  2285--2294, 2016.

\bibitem{rothman2013hitran2012}
L.~S. Rothman, I.~E. Gordon, Y.~Babikov, A.~Barbe, D.~C. Benner, P.~F. Bernath,
  M.~Birk, L.~Bizzocchi, V.~Boudon, L.~R. Brown \emph{et~al.}, ``{The HITRAN
  2012 Molecular Spectroscopic Database},'' \emph{{Journal of Quantitative
  Spectroscopy and Radiative Transfer}}, vol. 130, pp. 4--50, 2013.

\bibitem{srivastava2014dropout}
N.~Srivastava, G.~Hinton, A.~Krizhevsky, I.~Sutskever, and R.~Salakhutdinov,
  ``Dropout: a simple way to prevent neural networks from overfitting,''
  \emph{The Journal of Machine Learning Research}, vol.~15, no.~1, pp.
  1929--1958, 2014.

\bibitem{holland2008principal}
S.~M. Holland, ``Principal components analysis (PCA),'' \emph{Department of
  Geology, University of Georgia, Athens, GA}, pp. 30\,602--2501, 2008.

\bibitem{Allred_Partially}
\BIBentryALTinterwordspacing
C.~S. Allred. Partially correlated uniformly distributed random numbers.
  [Online]. Available:
  \url{https://medium.com/capital-one-tech/partially-correlated-uniformly-distributed-random-numbers-5ce82486b68a}
\BIBentrySTDinterwordspacing

\bibitem{Luaces2012}
\BIBentryALTinterwordspacing
O.~Luaces, J.~D{\'i}ez, J.~Barranquero, J.~J. del Coz, and A.~Bahamonde,
  ``Binary relevance efficacy for multilabel classification,'' \emph{Progress
  in Artificial Intelligence}, vol.~1, no.~4, pp. 303--313, Dec 2012. 
  \BIBentrySTDinterwordspacing

\bibitem{madden2009machine}
M.~G. Madden and T.~Howley, ``A machine learning application for classification
  of chemical spectra,'' in \emph{Applications and Innovations in Intelligent
  Systems XVI}.\hskip 1em plus 0.5em minus 0.4em\relax Springer, pp.
  77--90, 2009.

\bibitem{geladi1986partial}
P.~Geladi and B.~R. Kowalski, ``Partial least-squares regression: A tutorial,''
  \emph{Analytica Chimica Acta}, vol. 185, pp. 1--17, 1986.

\bibitem{elbassbasi2010classification}
M.~Elbassbasi, F.~Kzaiber, G.~Ragno, and A.~Oussama, ``Classification of raw
  milk by infrared spectroscopy ({FTIR}) and chemometric,'' \emph{Journal of
  Scientific Speculations and Research}, vol.~1, no.~2, pp. 28--33, 2010.

\bibitem{hirri2013classification}
A.~Hirri, M.~Bassbasi, and A.~Oussama, ``Classification and quality control of
  lubricating oils by infrared spectroscopy and chemometric,'' \emph{Int. J. Adv.
  Technol. Eng. Res.}, vol.~3, pp. 59--62, 2013.

\bibitem{hirri2016ftir}
A.~Hirri, M.~Bassbasi, S.~Platikanov, R.~Tauler, and A.~Oussama, ``FTIR
  spectroscopy and PLS-DA classification and prediction of four commercial
  grade virgin olive oils from Morocco,'' \emph{Food Analytical Methods},
  vol.~9, no.~4, pp. 974--981, 2016.

\end{thebibliography}
\end{document}